\documentclass{article}
\usepackage[eatrack]{log_2023}				

\usepackage{booktabs}						
\usepackage{multirow}						
\usepackage{amsfonts}						
\usepackage{graphicx}						
\usepackage{duckuments}						

\usepackage[
     backend=biber,
     style=numeric-comp,
     backref=true,
     natbib=true,
     sorting=none]{biblatex}
\AtEveryBibitem{%
  \clearfield{issn} 
  \clearfield{doi} 
  \clearfield{url}
  \clearfield{urldate}
  \clearfield{urlmonth}
  \clearfield{urlyear}
}

\usepackage{amsmath}
\usepackage{amssymb}
\usepackage{amsthm}
\newtheorem{theorem}{Theorem}
\newtheorem{lemma}{Lemma}

\newenvironment{hproof}{%
  \proof}{\endproof}

\usepackage[shortlabels]{enumitem}

\title[Over-Squashing in Riemannian Graph Neural Networks]{Over-Squashing in Riemannian Graph Neural Networks}

\author[J. Balla]{%
Julia Balla\\
Massachusetts Institute of Technology\\
jballa@mit.edu
}

\begin{document}

\maketitle

\begin{abstract}
Most graph neural networks (GNNs) are prone to the phenomenon of over-squashing in which node features become insensitive to information from distant nodes in the graph. Recent works have shown that the topology of the graph has the greatest impact on over-squashing, suggesting graph rewiring approaches as a suitable solution. In this work, we explore whether over-squashing can be mitigated through the embedding space of the GNN. In particular, we consider the generalization of Hyperbolic GNNs (HGNNs) to Riemannian manifolds of variable curvature in which the geometry of the embedding space is faithful to the graph's topology. We derive bounds on the sensitivity of the node features in these Riemannian GNNs as the number of layers increases, which yield promising theoretical and empirical results for alleviating over-squashing in graphs with negative curvature.
\end{abstract}

\section{Introduction}
\label{sec:introduction}

Graph Neural Networks (GNNs) have emerged as a powerful tool for modeling relational systems and learning on graph-structured data \cite{Scarselli,hamilton_inductive_2017, kipf2017semisupervised, pmlr-v70-gilmer17a}. Most GNN architectures rely on the \textit{message-passing} paradigm in which information is propagated along the edges of the graph, resulting in a class of Message Passing Neural Networks (MPNNs). However, due to an exponentially growing computational tree,  the compression of a quickly increasing amount of information into a fixed-size vector leads to informational \textit{over-squashing} \cite{alon2021on}. This phenomenon poses a significant challenge on long-range tasks with a large problem radius  since it obstructs the diffusion of information from distant nodes. 

The over-squashing problem has been analyzed through various lenses such as graph curvature \cite{topping_understanding_2022}, information theory \cite{banerjee_oversquashing_2022}, and effective resistance \cite{black_understanding_2023}, each suggesting a corresponding approach to mitigate the issue by rewiring the graph. Along with several other works, this line of reasoning has resulted in a ``zoo'' of proposed graph rewiring techniques for over-squashing \cite{abboud2022shortest, arnaiz-rodriguez2022diffwire,deac2022expander,karhadkar2023fosr}. Recent work has unified the spatial and spectral techniques under a common framework and justified their efficacy by demonstrating that graph topology plays the biggest role in alleviating over-squashing as opposed to MPNN properties such as width or depth \cite{di_giovanni_over-squashing_2023} .

One potential drawback of many spatial graph rewiring techniques is the distortion of structural information that may be relevant to the learning task. Instead of altering the graph topology, we thus consider augmentations to the MPNN architecture that would make it topology-aware. Specifically, we explore the effects of changing the embedding space of the GNN. The hypothesis behind our approach is that by embedding the negatively curved sections of the graph in hyperbolic space, there would be less information lost at each layer due to the increased representational capacity. However,  hyperbolic space is a poor inductive bias for graphs with significant positive curvature, where spherical space would be more suitable. 
Therefore, we consider a GNN that embeds graphs in Riemannian manifolds of variable curvature.

We study the over-squashing phenomenon in one such model by generalizing the Hyperbolic GNN (HGNN) architecture \cite{liu_hyperbolic_2019} to Riemannian GNNs (RGNNs). Assuming that there exists a Riemannian manifold where the geometry matches that of the input graph, RGNNs are in principle able to embed the graph in this manifold. While the RGNN architecture is not immediately computationally tractable in its most general form, it provides a means to derive a best-case theoretical result on over-squashing. We derive a bound on the Jacobian of the node features in a RGNN and show that it relies on the global curvature properties of the embedding space. Based on this bound, we heuristically and empirically demonstrate that our model addresses cases where the graph's curvature is predominantly negative everywhere (e.g. tree-like graphs). We also identify pathological cases where our model may fail on manifolds with both positive and negative curvature. Finally, we propose concrete next steps to complete our theoretical analysis that would justify step (2) in the argument above and motivate the development of tractable methods that approximate general Riemannian GNNs.

\section{Riemannian GNNs}
\label{sec:riemannian_gnns}

For a primer on the Riemannian geometry notions used throughout the following sections, we refer the reader to Appendix \ref{appendix:riemannian}. We define GNNs that embed node representations in a Riemannian space that is faithful to the input graph's topology. Crucially, we assume that we are given an ``optimal'' Riemannian manifold $(\mathcal{M}, g)$ and that the GNN has access to the distance, exponential map, and logarithmic map functions as differentiable operations. While finding an optimal Riemannian manifold of variable curvature is challenging in practice, there exist methods for its approximation \cite{xiong_pseudo-riemannian_2022, gu2018learning, giovanni2022heterogeneous, Cruceru2020ComputationallyTR, lopez_symmetric_2021}. For the purposes of our analysis, we assume this approximation of $(\mathcal{M}, g)$ is exact.

To generalize the Euclidean GNNs to Riemannian manifolds, \citet{liu_hyperbolic_2019} build upon Hyperbolic Neural Networks (HNNs) \cite{ganea_hyperbolic_2018}. Since there is no well-defined notion of vector space structure in Riemannian space, the main idea is to leverage the exponential and logarithmic maps to perform node feature transformation and neighborhood aggregation functions as Euclidean operations in the tangent space $\mathcal{T}_{\mathbf{p}}\mathcal{M}$ of some chosen point $\mathbf{p} \in \mathcal{M}$. In particular, the node update rule is given by
\begin{equation}\label{eq:riemannian_gcn}
    \mathbf{x}_i^{(\ell+1)}=\sigma\left(\exp _{\mathbf{p}}\left(\sum_{j \in \mathcal{N}(i)} \tilde{\mathbf{A}}_{ij} \mathbf{W}^{(\ell)} \log _{\mathbf{p}}\left(\mathbf{x}_j^{(\ell)}\right)\right)\right)
\end{equation}
where $\tilde{\mathbf{A}} = \mathbf{D^{-\frac{1}{2}}}(\mathbf{A} + \mathbf{I})\mathbf{D^{-\frac{1}{2}}}$ is the normalized adjacency matrix with self-loops, $\mathcal{N}(i)$ is the set of in-neighbors of node $i$, $\mathbf{W}^{(\ell)}$ is the matrix of trainable parameters at layer $\ell$, and $\sigma$ is a chosen activation function. Note that in the case of the Euclidean manifold, operating in the tangent space of the origin by setting $\mathbf{p} = \mathbf{o}$ recovers a vanilla GNN. Since hyperbolic manifolds fall under the class of manifolds that have a pole $\mathbf{o}$ (i.e., $\exp_\mathbf{o}: \mathcal{T}_{\mathbf{o}}\mathcal{M} \to \mathcal{M}$ is a diffeomorphism \cite{itoh1980}), \citet{liu_hyperbolic_2019} choose $\mathbf{p} = \mathbf{o}$ across all nodes and layers for HGNNs. However, general Riemannian manifolds do not have a pole, so we let $\mathbf{p}=p(i, \ell )\in \mathcal{M}$ for an arbitrary function $p$ that depends on the current node and/or the layer $\ell$. We leave the selection of an optimal function $p$ as future work. We also ensure that the exponential and logarithmic maps are differentiable by restricting $\left \|\sum_{j \in \mathcal{N}(i)} \tilde{\mathbf{A}}_{ij} \mathbf{W}^{(\ell)} \log _{\mathbf{p}}\left(\mathbf{x}_j^{(\ell)}\right)\right\|_2$ to fall within the injectivity radius of $\mathbf{p}$.

\section{Sensitivity Analysis}
\label{sec:sensitivity}

Following the methodology in \cite{di_giovanni_over-squashing_2023}, we assess the over-squashing effect in RGNNs by deriving a bound on the norm of the Jacobian of node features after $\ell$ layers. 
Since this involves bounding the differentials of the exponential and logarithmic maps, we first derive the following lemma.
\begin{lemma}\label{lemma:differentials}
    Consider a RGNN as in equation \eqref{eq:riemannian_gcn} with Riemannian manifold $(\mathcal{M}, g)$ with bounded sectional curvature $k \leq \kappa_\mathbf{p}(\mathbf{u}, \mathbf{v}) \leq K$ for all $\mathbf{p} \in \mathcal{M}$ and $\mathbf{u},\mathbf{v} \in \mathcal{T}_\mathbf{p}\mathcal{M}$. Let $Df$ denote the differential of a map $f$. Then for $\exp_\mathbf{p}$ and $\log_\mathbf{p}$ in \eqref{eq:riemannian_gcn} and $i \in V$ we have
    \begin{align*}
        &\left\|D \exp_\mathbf{p} \right\|_2  \left\|D \log_\mathbf{p} \right\|_2 \leq 
        \begin{cases}
            \frac{\operatorname{sinh}\left(\sqrt{-k}r_{i, \exp}\right)}{\sqrt{-k}r_{i, \exp}} & k < K \leq 0\\
            \frac{\operatorname{sinh}\left(\sqrt{-k}r_{i, \exp}\right)\operatorname{sin}\left(\sqrt{K}r_{i, \log}\right)}{\sqrt{-kK}r_{i, \exp}r_{j, \log}}  & k < 0 < K\\
            \frac{\operatorname{sin}\left(\sqrt{K}r_{i, \log}\right)}{\sqrt{K}r_{i, \log}} & 0 \leq k < K\\
            1 & k = K = 0
        \end{cases}
        =: \beta_i(k,K) 
    \end{align*}
    where $r_{i, \exp} = \sup\limits_{\ell} \left\| \sum_{z \in \mathcal{N}(i)} \tilde{\mathbf{A}}_{iz} \mathbf{W}^{(\ell)} \log _{\mathbf{p}}\left(\mathbf{x}_z^{(\ell)}\right) \right\|_2$ denotes the maximum radius around $\mathbf{p}$ for the exponential map
    and $r_{i, \log} = \sup\limits_{z, \ell} \|\mathbf{x}_z^{(\ell)}\|_g$ is the maximum radius for the logarithmic map.
\end{lemma}

The proof for the above lemma relies on a well-known sectional curvature comparison result in differential geometry and can be found in Appendix \ref{appendix:differentials}. We use this lemma to derive a bound on the sensitivity of node features.

\begin{theorem} \label{thm:main}
    Under the same assumptions as in Lemma \ref{lemma:differentials}, if
    $c_\sigma$ is the Lipschitz constant of the nonlinearity $\sigma$ and $w \geq \left\| \mathbf{W}^{(l)} \right\|_2$ is an upper bound on the spectral norm of all weight matrices,
    then for $i,j \in V$
    $$
    \left\| \frac{\partial \mathbf{x}_i^{(\ell)}}{\partial \mathbf{x}_j^{(0)}} \right\|_2 \leq c_\sigma^{\ell} w^{\ell}
        \beta_i(k,K)^{\ell} \left(\tilde{\mathbf{A}}^{\ell}\right)_{ij}
    $$
    where $\beta_i(k,K)$ is a bound on the sensitivity of the exponential and logarithmic maps as defined in Lemma \ref{lemma:differentials}.
\end{theorem}
The proof uses induction over the number of layers $\ell$ and is provided in Appendix \ref{appendix:theorem}. Note that this bound has the same form as in \cite{di_giovanni_over-squashing_2023} for classical GNNs, and in fact is equivalent for Euclidean space (i.e., $k=K=0$). To show that the RGNN is able to compensate for the information bottlenecks arising from taking powers of the adjacency matrix, \textbf{it remains to demonstrate that the growth (decay) of $\beta_i(k,K)^{\ell}$ is able to mitigate the decay (growth) of $(\tilde{\mathbf{A}}^{\ell})_{ij}$ as $\ell$ increases.} In Appendix \ref{appendix:hyperbolic_example}, we demonstrate that this property holds for the pathological example of negative curvature mentioned in \cite{topping_understanding_2022}. While a formal analysis of the variable curvature case is left as future work, we provide a heuristic argument based on the magnitude of $k$ and $K$. 
\begin{hproof}
    Assume that $|r_{i, \exp}|$ and  $|r_{i, \log}|$ do not grow very small or large as $\ell$ increases. If $k <0$ and $|k| << |K|$, $\beta_i(k,K)$ is dominated by the term $\frac{\operatorname{sinh}\left(\sqrt{-k}r_{i, \exp}\right)}{\sqrt{-k}r_{i, \exp}}$ which increases as $k$ grows more negative. Therefore, $\beta_i(k,K)^\ell$ grows large as $\ell$ increases and thus helps to alleviate over-squashing. On the other hand, if $K > 0$ and $|K| >> |k|$, $\beta_i(k,K)$ is dominated by the term $\frac{\operatorname{sin}\left(\sqrt{k}r_{i, \log}\right)}{\sqrt{k}r_{i, \log}}$ which decreases (albeit non-monotonically) as $k$ grows more positive. Then $\beta_i(k,K)^\ell$ grows small as $\ell$ increases and instead hinders the flow of information from $j$ to $i$. This behavior is not problematic since graphs with positive curvature (corresponding to cycles) would have already exchanged overlapping information in the earlier layers. However, an issue may arise in the case when $k < 0 < K$ and $|k| << |K|$ for which $\beta_i(k,K)^\ell$ grows small despite the existence of very negatively curved sections of the graph. 
\end{hproof}

This argument highlights a limitation of the result in Theorem \ref{thm:main} in that the bound only depends on global sectional curvature bounds $k$ and $K$. Therefore, $\beta_i(k,K)$ does not target the sensitivity of specific node pairs induced by $(\tilde{\mathbf{A}}^{\ell})_{ij}$. Note that if we let $\mathbf{p}=p(i, \ell, \mathbf{x}_i^{(\ell)})\in \mathcal{M}$ be a function of the current node feature,  the neighboring feature aggregation would intuitively depend on the local curvature at $\mathbf{x}_i^{(\ell)} \in \mathcal{M}$. However, this would significantly increase the complexity of the Riemannian GNN model and hence the Jacobian sensitivity derivation.

\section{Empirical Results}
Given that the special case of Hyperbolic GNNs is well-defined and computationally tractable, we compare the empirical sensitivity of node features in Hyperbolic Graph Convolutional Networks (HGCNs) \cite{chami_hyperbolic_2019} to Euclidean GCNs. We use the link prediction benchmark datasets (as well as the model hyperparameters) provided in \cite{chami_hyperbolic_2019}: citation networks (Cora \cite{Sen_Namata_Bilgic_Getoor_Galligher_Eliassi-Rad_2008} and PubMed \cite{Namata2012QuerydrivenAS}), disease propogation trees (Disease), and flight networks (Airport). The Gromov $\delta$-hyperbolicity value of each dataset is reported in Figure 1, where lower $\delta$ is more hyperbolic. Since over-squashing is more severe for deeper GNNs, we evaluate GCNs and HGCNs (specifically the Poincar\'e model) of depth $6$. We then consider 100 randomly sampled pairs of nodes that are distance $6$ apart and take the average of the norm of their Jacobians, $\frac{1}{100} \sum_{(i, j)} \left\| \frac{\partial \mathbf{x}_i^{(6)}}{\partial \mathbf{x}_j^{(0)}} \right\|_2$. As shown in Figure 1, for three of the four datasets, both the average and maximum sensitivity in the sample are greater in HGCNs than in GCNs at each epoch. For PubMed, while the average sensitivities are roughly equal, the maximum is still always greater for HGCNs, which is consistent with our upper bound in Theorem 1. The results hold even for Cora, which has a higher hyperbolicity value. This suggests that hyperbolic embeddings may be sufficient for alleviating over-squashing even in non-hyperbolic graphs, as the distortion of positively curved regions could be compensated for by the increased sensitivity between node pairs in those regions.

We limit our empirical analysis to the special case of hyperbolic manifolds since the implementation of Riemannian GNNs as defined in \eqref{eq:riemannian_gcn} is not immediately feasible. First of all, it is not obvious how the reference point $\mathbf{p}$ should be defined at any given node. Moreover, our analysis assumes that we are given an optimal manifold in which the GNN should embed the graph. As described in Appendix \ref{appendix:approximations}, it is not trivial to obtain the exact manifold for heterogeneous embedding spaces. However, there exist several methods for approximating these manifolds \cite{xiong_pseudo-riemannian_2022, gu2018learning, giovanni2022heterogeneous, Cruceru2020ComputationallyTR, lopez_symmetric_2021}, many of which have desirable properties such as well-defined origin points for $\mathbf{p}$. We leave an empirical study of over-squashing in RGNNs built on these approximations as future work.

\begin{figure}[t]
    \centering
    \includegraphics[width=1\linewidth]{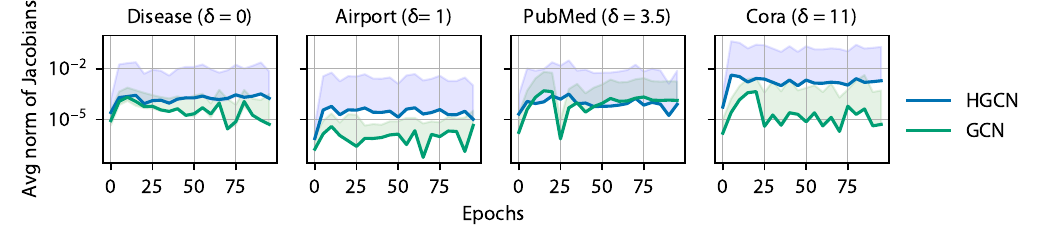}
    \caption{Sensitivity of node representations at layer $6$ with respect to the input node features. The solid line denotes the average norm of the Jacobians for a random sample of 100 node pairs that are $6$ hops apart. The shaded regions indicate the intervals between the minimum and maximum norm values (where the minimums tend to be very close to the average). We also include the hyperbolicity values $\delta$ of each dataset provided in \cite{chami_hyperbolic_2019}.}
    \label{fig:enter-label}
\end{figure}

\section{Discussion}
\label{sec:discussion}

We derive a bound on the Jacobian of node features in a Riemannian GNN. The bound contains a global curvature-dependent term $\beta_i(k,K)$ that grows exponentially with the number of layers $\ell$ when the embedding space has a minimum sectional curvature which is very negative and decays exponentially when the space has very positive maximum curvature. Since information bottlenecks have been linked to negative curvature on graphs, the exponential growth when $k<0$ is a promising result for mitigating over-squashing. 

Despite the heuristic argument provided in section 3 and promising empirical results for Hyperbolic GNNs in section 4, we do not formally prove that $\beta_i(k,K)$ compensates for the exponential decay of $(\tilde{\mathbf{A}}^{\ell})_{ij}$ as $
\ell$ increases without hindering overall model performance. One potential approach to deriving the relationship between the two terms could involve connecting the $\beta_i(k,K)$ term to edge-based Ricci curvature and utilizing the results in \cite{topping_understanding_2022}. Using the intuition that the Ricci curvature can be considered as an ``average'' over sectional curvatures, it may be possible to define a notion of sectional curvature on a graph (e.g. the one proposed by \citet{gu2018learning}) such that the Balanced Forman curvature in \cite{topping_understanding_2022} is an average of curvatures assigned to triangles of nodes. This connection may allow one to quantify how $(\tilde{\mathbf{A}}^{\ell})_{ij}$ is affected by both local and global sectional curvature. 
Additionally, due to the Riemannian GNN's dependence on global curvature properties, the model may end up in a pathological scenario when the decay in sensitivity from maximum positive curvature outweighs the growth from the minimum negative curvature. This may call for the introduction of local curvature information into the architecture such that the neighbor aggregation at node $i$ explicitly depends on the curvature near $i$. It may also be possible to localize the sensitivity bounds by constraining the manifold to have \textit{locally} bounded sectional curvature everywhere.

Finally, while the Riemannian GNN is useful for the theoretical over-squashing analysis, implementing the proposed architecture comes with several challenges. It would be exciting to see the development of models that can more closely approximate Riemannian GNNs while maintaining tractability. For instance, it may be possible to apply the deep Riemannian manifold learning in \cite{lou_deep_nodate} such that the optimal manifold $(\mathcal{M}, g)$ is parameterized as a neural network itself. We hope that the insights gained from our theoretical results will inspire future work in the development of practical architectures that leverage these findings.

\printbibliography

\appendix
\section{Riemannian Geometry} \label{appendix:riemannian}
We first introduce some preliminary notation and concepts in Riemannian geometry. We refer the reader \cite{petersen_riemannian_2016} for a more detailed discussion of these concepts.

A Riemannian manifold $(\mathcal{M}, g)$ is a smooth manifold equipped with a \textit{Riemannian metric} $g_\mathbf{x}: \mathcal{T}_\mathbf{x}\mathcal{M} \times \mathcal{T}_\mathbf{x}\mathcal{M} \to \mathbb{R}$ where $\mathcal{T}_\mathbf{x}\mathcal{M}$ is the tangent space at the point $\mathbf{x} \in \mathcal{M}$. The Riemannian metric is a local inner product that varies smoothly with $\mathbf{x}$ and allows us to define the geometric properties of a space such as length, angle, and area. For instance, $g$ induces a norm  $\|\mathbf{v}\|_g = \sqrt{g_{\mathbf{x}}(\mathbf{v}, \mathbf{v})}$ for any $v \in \mathcal{T}_\mathbf{x}\mathcal{M}$.

\subsection{Geodesics} 
The Riemannian metric also gives rise to a notion of distance. For a curve $\gamma: [0, T] \to \mathcal{M}$, the length of $\gamma$ is given by $L(\gamma) = \int_0^T \| \gamma'(t) \|_g dt$.  Thus, for two points $\mathbf{x}, \mathbf{y} \in \mathcal{M}$, the distance is defined as $d_g(\mathbf{x}, \mathbf{y}) = \inf L(\gamma)$ where $\gamma$ is any curve such that $\gamma(0) = \mathbf{x}$ and $\gamma(T) = \mathbf{y}$. A \textit{geodesic} is a curve that minimizes this length.

\subsection{Exponential and Logarithmic Map} 
For each point $\mathbf{x} \in \mathcal{M}$ and velocity vector $\mathbf{v} \in \mathcal{T}_\mathbf{x}\mathcal{M}$, there exists a unique geodesic $\gamma: [0,1] \to \mathcal{M}$ where $\gamma(0)=\mathbf{x}$ and $\gamma'(0)=\mathbf{v}$. The \textit{exponential map} $\exp_\mathbf{x}: \mathcal{T}_\mathbf{x}\mathcal{M} \to \mathcal{M}$ is defined as $\exp_\mathbf{x}(\mathbf{v}) = \gamma(1)$. Its local inverse is called the \textit{logarithm map}, $\log_\mathbf{x}(\mathbf{v})$. Note that the distance between two points $\mathbf{x}, \mathbf{y} \in \mathcal{M}$ can be represented as  $d_g(\mathbf{x}, \mathbf{y}) = \|\log_\mathbf{x}(\mathbf{y})\|_g$. 

Manifolds where the exponential map is defined on the whole tangent space $\mathcal{T}_\mathbf{x}\mathcal{M}$ are called \textit{geodesically complete}. However, geodesic completeness does not guarantee that the exponential map is a global diffeomorphism (i.e. a differentiable bijective map with a differentiable inverse). The radius of the largest ball about the origin in $\mathcal{T}_\mathbf{x}\mathcal{M}$ that can be mapped diffeomorphically via the exponential map is called the \textit{injectivity radius} of $\mathcal{M}$ at $\mathbf{x}$.

\subsection{Curvature} 
For each point $\mathbf{x} \in \mathcal{M}$ and pair of linearly independent tangent vectors $\mathbf{u}, \mathbf{v} \in \mathcal{T}_\mathbf{x}\mathcal{M}$, the \textit{sectional curvature} $\kappa_\mathbf{x}(\mathbf{u}, \mathbf{v})$ at $\mathbf{x}$ is defined as the \textit{Gaussian curvature} of the  two-dimensional surface obtained by exponentiating a plane spanned by $\mathbf{u}$ and $\mathbf{v}$ at $\mathbf{x}$. The Gaussian curvature of a surface is given by the product of the principal curvatures. Riemannian manifolds of constant sectional curvature $\kappa$ are called \textit{space forms}, the most common examples being spherical space ($\kappa > 0$), Euclidean space ($\kappa = 0$), and hyperbolic space ($\kappa < 0$). Another form of curvature on a Riemannian manifold is \textit{Ricci curvature}, which is a symmetric bilinear form determining the geodesic dispersion at nearby points. The Ricci curvature of a tangent vector $\mathbf{v}$ at $\mathbf{p}$ is the average of the sectional curvature over all tangent planes containing $\mathbf{v}$.

Several works have also introduced discrete notions of sectional and Ricci curvature on graphs. \citet{gu2018learning} introduced a discrete notion of sectional curvature for learning product manifolds of mixed curvatures for graph embeddings. \citet{Forman2003} and \citet{Ollivier2007,Ollivier2007RicciCO} proposed edge-based curvature that could recover certain properties of the Ricci curvature on manifolds. \citet{topping_understanding_2022} used a novel formulation of Ricci curvature to show that over-squashing in GNNs is related to the existence of edges with high negative curvature.

\subsection{Riemannian Manifolds for Graph Embeddings} \label{appendix:approximations}

There has been a surge in the development of algorithms that represent graphs as sets of node embeddings in hyperbolic and spherical space due to their favorable geometric inductive biases \cite{wilson,nickel_poincare_2017,chamberlain2017neural,liu_sphereface_2017,sala_representation_2018}. These space forms are well defined and offer closed-form expressions for geometric operations such as the exponential and logarithmic map, making them suitable for optimization in these spaces. 

However, space forms individually may not capture all of the geometric properties of a given graph. On the other hand, heterogeneous manifolds of variable curvature lack computational tractability. Several works have instead embedded graphs in manifolds of mixed curvature by taking Cartesian products of homogenous model spaces \cite{gu2018learning}, adding heterogeneous dimensions to homogenous spaces \cite{giovanni2022heterogeneous}, or limiting the embedding space to certain classes of manifolds \cite{Cruceru2020ComputationallyTR, lopez_symmetric_2021}. An exciting direction for learnable Riemannian manifolds has been proposed by \citet{lou_deep_nodate}, where the metric is parametrized by a deep neural network.

\section{Example: Sensitivity for a Binary Tree in Hyperbolic Space} \label{appendix:hyperbolic_example}

Suppose that nodes $i$ and $j$ are distance $\ell + 1$ apart and that the receptive field of node $i$ is a binary tree in a RGNN given a manifold with constant negative sectional curvature $k<0$ (i.e. a Hyperbolic GNN). Then $(\tilde{\mathbf{A}}^{\ell})_{ij} = 2^{-1}3^{-\ell}$ and, by Theorem \ref{thm:main}, 
\begin{equation*}
    \beta_i(k,k)^\ell = \left(\frac{\operatorname{sinh}\left(\sqrt{-k}r_{i, \exp}\right)}{\sqrt{-k}r_{i, \exp}}\right)^\ell 
\end{equation*}
Therefore, $\beta_i(k,k)^\ell > (\tilde{\mathbf{A}}^{\ell})_{ij}$ when
\begin{align*}
    \left(\frac{\operatorname{sinh}\left(\sqrt{-k}r_{i, \exp}\right)}{\sqrt{-k}r_{i, \exp}}\right)^\ell &> \frac{1}{3^{\ell}} > \frac{1}{2\cdot3^{\ell}}\\
    \frac{\operatorname{sinh}\left(\sqrt{-k}r_{i, \exp}\right)}{\sqrt{-k}r_{i, \exp}} &> \frac{1}{3}.
\end{align*}
This example suggests that over-squashing is indeed less severe in HGNNs on graphs exhibiting negative curvature. 

\section{Proof of Lemma \ref{lemma:differentials}} \label{appendix:differentials}

We first note a comparison lemma from chapter 6.2 in \cite{petersen_riemannian_2016} that yields bounds on the differential of the exponential and logarithmic maps.

\begin{lemma} \label{lemma:peterson}
Assume that $(\mathcal{M}, g)$ satisfies $k \leq K_\mathbf{x}(\mathbf{u}, \mathbf{v}) \leq K$ for all $\mathbf{x} \in \mathcal{M}$ and $\mathbf{u},\mathbf{v} \in \mathcal{T}_\mathbf{x}\mathcal{M}$. Let $Df$ denote the differential of a map $f$. Then for the exponential and logarithmic map at $\mathbf{x}$ and for a radius $r$ around $\mathbf{x}$ we have 
\begin{equation*}
    \begin{gathered}
    \left\|D \exp_\mathbf{x}\right\|_2 \leq \max \left\{1, \frac{\operatorname{sn}_k(r)}{r}\right\},\\
    \left\|D \log_\mathbf{x}\right\|_2 \leq \min \left\{1, \frac{\operatorname{sn}_K(r)}{r}\right\}
    \end{gathered}
\end{equation*}
where $\operatorname{sn}_\kappa(\cdot)$ is the generalized sine function given sectional curvature $\kappa$
\begin{equation*}
    \mathrm{sn}_\kappa(r):= \begin{cases}\frac{\sin (\sqrt{\kappa} r)}{\sqrt{\kappa}} & \text { if } \kappa>0 \\ r & \text { if } \kappa=0 \\ \frac{\sinh (\sqrt{-\kappa} r)}{\sqrt{-\kappa}} & \text { if } \kappa<0\end{cases}.
\end{equation*}
\end{lemma}

We use the above lemma to derive a bound on the product of norms of the exponential and logarithmic maps in equation \eqref{eq:riemannian_gcn} as stated in Lemma \ref{lemma:differentials}.

\begin{proof}
    Let $r_{j, \exp} = \sup\limits_{\ell} \left\| \sum_{z \in \mathcal{N}(j)} \tilde{\mathbf{A}}_{jz} \mathbf{W}^{(\ell)} \log _{\mathbf{p}}\left(\mathbf{x}_z^{(\ell)}\right) \right\|_2$ denote the maximum radius around $\mathbf{x}$ for the exponential map and $r_{j, \log} = \sup\limits_{\ell} \|\mathbf{x}_z^{(\ell)}\|_g$ denote the maximum radius for the logarithmic map given equation \eqref{eq:riemannian_gcn}. Applying Lemma \ref{lemma:peterson}, there are three possible cases for the bounds $k$ and $K$:
    
    \underline{Case 1:} $k < K \leq 0$. We then have
    \begin{align*}
        \left\|D \exp_\mathbf{p}\right\|_2  \left\|D \log_\mathbf{p} \right\|_2  \leq \max \left\{1, \frac{\operatorname{sinh}\left(\sqrt{-k}r_{j, \exp}\right)}{\sqrt{-k}r_{j, \exp}}\right\} \cdot \max_{z \in \mathcal{N}(j)}
     \min \left\{1, \frac{\operatorname{sinh}\left(\sqrt{-K}r_{j, \log}\right)}{\sqrt{-K}r_{j, \log}}\right\}.  
    \end{align*}
    Since $\frac{\sinh(x)}{x} > 1$ for all $x \neq 0$, we obtain the bound
    \begin{align*}
        \left\|D \exp_\mathbf{p}\right\|_2  \left\|D \log_\mathbf{p} \right\|_2  \leq \frac{\operatorname{sinh}\left(\sqrt{-k}r_{j, \exp}\right)}{\sqrt{-k}r_{j, \exp}}.
    \end{align*}
    \underline{Case 2:} $k < 0 < K$. We then have
    \begin{align*}
        \left\|D \exp_\mathbf{p} \right\|_2  \left\|D \log_\mathbf{p} \right\|_2  \leq \frac{\operatorname{sinh}\left(\sqrt{-k}r_{j, \exp}\right)}{\sqrt{-k}r_{j, \exp}} \cdot \max_{z \in \mathcal{N}(j)}
     \min \left\{1, \frac{\operatorname{sin}\left(\sqrt{K}r_{j, \log}\right)}{\sqrt{K}r_{j, \log}}\right\}.  
    \end{align*}
    Since $\frac{\sin(x)}{x} < 1$ for all $x \neq 0$, we obtain the bound
    \begin{align*}
        \left\|D \exp_\mathbf{p}\right\|_2  \left\|D \log_\mathbf{p} \right\|_2  &\leq \frac{\operatorname{sinh}\left(\sqrt{-k}r_{j, \exp}\right)}{\sqrt{-k}r_{j, \exp}} \cdot \max_{z \in \mathcal{N}(j)} \frac{\operatorname{sin}\left(\sqrt{K}r_{j, \log}\right)}{\sqrt{K}r_{j, \log}}.
    \end{align*}
    \underline{Case 3:} $ 0 \leq k < K$. We then have
    \begin{align*}
        \left\|D \exp_\mathbf{p} \right\|_2  \left\|D \log_\mathbf{p} \right\|_2  &\leq \max \left\{1, \frac{\operatorname{sin}\left(\sqrt{k}r_{j, \exp}\right)}{\sqrt{k}r_{j, \exp}}\right\} \cdot \max_{z \in \mathcal{N}(j)}
     \min \left\{1, \frac{\operatorname{sin}\left(\sqrt{K}r_{j, \log}\right)}{\sqrt{K}r_{j, \log}}\right\}\\ 
     &= \max_{z \in \mathcal{N}(j)} \frac{\operatorname{sin}\left(\sqrt{K}r_{j, \log}\right)}{\sqrt{K}r_{j, \log}}.\\ 
    \end{align*}
    \underline{Case 4:} $0 = k = K$. Then we have 
    \begin{align*}
        \left\|D \exp_\mathbf{p} \right\|_2  \left\|D \log_\mathbf{p} \right\|_2  &\leq \max \left\{1, \frac{r_{j, \exp}}{r_{j, \exp}}\right\} \cdot \max_{z \in \mathcal{N}(j)}
     \min \left\{1, \frac{r_{j, \log}}{r_{j, \log}}\right\} = 1.
    \end{align*}
    Combining all of the cases above, we obtain the bound
    \begin{align*}
        \left\|D \exp_\mathbf{p} \right\|_2  \left\|D \log_\mathbf{p}\right\| &\leq  \begin{cases}
            \frac{\operatorname{sinh}\left(\sqrt{-k}r_{j, \exp}\right)}{\sqrt{-k}r_{j, \exp}} & k < K \leq 0\\
            \frac{\operatorname{sinh}\left(\sqrt{-k}r_{j, \exp}\right)}{\sqrt{-k}r_{j, \exp}} \cdot \max_{z \in \mathcal{N}(j)} \frac{\operatorname{sin}\left(\sqrt{K}r_{j, \log}\right)}{\sqrt{K}r_{j, \log}} & k < 0 < K\\
            \max_{z \in \mathcal{N}(j)} \frac{\operatorname{sin}\left(\sqrt{K}r_{j, \log}\right)}{\sqrt{K}r_{j, \log}} & 0 \leq k < K\\
            1 & k = K = 0
        \end{cases} \\
        &= \beta_j(k,K).
    \end{align*}
\end{proof}

\section{Proof of Theorem \ref{thm:main}} \label{appendix:theorem}

\begin{proof}
    We prove the bound by induction on the number of layers $\ell$. For the base case of $\ell=1$, we have
    \begin{align*}
        \left\| \frac{\partial \mathbf{x}_i^{(1)}}{\partial \mathbf{x}_j^{(0)}} \right\|_2 &= \left\| \frac{\partial}{\partial \mathbf{x}_j^{(0)}} \left[\sigma\left(\exp _{\mathbf{p}}\left(\sum_{z \in \mathcal{N}(i)} \tilde{\mathbf{A}}_{iz} \mathbf{W}^{(0)} \log _{\mathbf{p}}\left(\mathbf{x}_z^{(0)}\right)\right)\right) \right]\right\|_2 \\
        &\leq c_\sigma \left\|D \exp_\mathbf{p} \right\|_2  \left\| \mathbf{W}^{(0)} \right\|_2 \left\|D \log_\mathbf{p} \right\|_2 \sum_{z \in \mathcal{N}(i)} \tilde{\mathbf{A}}_{iz}  \left\|\frac{\partial \mathbf{x}_z^{(0)}}{\partial \mathbf{x}_j^{(0)}}\right\|_2\\
        &\leq c_\sigma w \left\|D \exp_\mathbf{p} \right\|_2  \left\|D \log_\mathbf{p} \right\|_2  \tilde{\mathbf{A}}_{ij}  \left\|\frac{\partial \mathbf{x}_j^{(0)}}{\partial \mathbf{x}_j^{(0)}}\right\|_2\\
        &= c_\sigma w \tilde{\mathbf{A}}_{ij}  \left\|D \exp_\mathbf{p} \right\|_2  \left\|D \log_\mathbf{p} \right\|_2.  
    \end{align*}
    If we let $\beta_i(k, K)$ be the bound on $\left\|D \exp_\mathbf{p} \right\|_2  \left\|D \log_\mathbf{p}\right\|$ defined in Lemma \ref{lemma:peterson}, 
    the norm of the Jacobian in the base case (i.e. $\ell=1$) is bounded by
    \begin{align*}
        \left\| \frac{\partial \mathbf{x}_i^{(1)}}{\partial \mathbf{x}_j^{(0)}} \right\|_2 
        &\leq c_\sigma w \beta_i(k, K) \tilde{\mathbf{A}}_{ij}.
    \end{align*}
    We now assume the bound to be satisfied for $\ell$ layers and use induction to show that it holds for $\ell+1$.
    \begin{align*}
        \left\| \frac{\partial \mathbf{x}_i^{(\ell+1)}}{\partial \mathbf{x}_j^{(0)}} \right\|_2 &= \left\| \frac{\partial}{\partial \mathbf{x}_j^{(0)}} \left[\sigma\left(\exp_{\mathbf{p}}\left(\sum_{z \in \mathcal{N}(i)} \tilde{\mathbf{A}}_{iz} \mathbf{W}^{(\ell)} \log _{\mathbf{p}}\left(\mathbf{x}_z^{(\ell)}\right)\right)\right) \right]\right\|_2 \\
        &\leq c_\sigma w 
        \left\|D \exp_\mathbf{p} \right\|_2 \left\|D \log_\mathbf{p} \right\|_2 \sum_{z \in \mathcal{N}(i)} \tilde{\mathbf{A}}_{iz}   \left\|\frac{\partial \mathbf{x}_z^{(\ell)}}{\partial \mathbf{x}_j^{(0)}}\right\|_2\\
        &\leq c_\sigma w 
        \beta_i(k,K) \sum_{z \in \mathcal{N}(i)} \tilde{\mathbf{A}}_{iz} \left[c_\sigma^\ell w^\ell  \beta_i(k, K)^\ell \left(\tilde{\mathbf{A}}^\ell\right)_{zj} \right]\\
        &= c_\sigma^{\ell+1} w^{\ell+1}
        \beta_i(k,K)^{\ell+1} \sum_{z \in \mathcal{N}(i)} \tilde{\mathbf{A}}_{iz} \left(\tilde{\mathbf{A}}^\ell\right)_{zj} \\
        &= c_\sigma^{\ell+1} w^{\ell+1}
        \beta_i(k,K)^{\ell+1} \left(\tilde{\mathbf{A}}^{\ell+1}\right)_{ij}. \\
    \end{align*}
    
\end{proof}

\end{document}